\documentclass[1p,review]{elsarticle}
\usepackage{hyperref}
\usepackage[utf8]{inputenc} 
\usepackage[T1]{fontenc}    
\usepackage[tbtags,fixamsmath]{mathtools}
\usepackage{subfigure}
\usepackage{booktabs,multirow,multicol,ctable}

\usepackage{color}
\usepackage{xcolor}
\usepackage{soul}

\newif\ifrevfinal
\revfinalfalse
\def\rev[#1][#2]{\ifrevfinal #2 \else {\color{blue} \sout{#1}} {\bf \color{red} #2} \fi}

\usepackage{amsmath,amssymb,amsbsy,xspace}
\makeatletter
\DeclareRobustCommand\onedot{\futurelet\@let@token\@onedot}
\def\@onedot{\ifx\@let@token.\else.\null\fi\xspace}

\def\ie{\emph{i.e}\onedot}

\def\be{\begin{equation}}
\def\ee{\end{equation}}
\def\bea{\begin{eqnarray}}
\def\eea{\end{eqnarray}}

\makeatother

\newcommand{\myparagraphwithspace}[1]{\vspace{6pt}\noindent{\bf #1}}

\journal{Image and Vision Computing}
\begin{document}

\begin{frontmatter}

\title{How Robust are Discriminatively Trained Zero-Shot Learning Models?}

\author[hacettepeaddress]{Mehmet Kerim Yucel}
\ead{mkerimyucel@hacettepe.edu.tr}

\author[odtuaddress]{Ramazan Gokberk Cinbis}
\ead{gcinbis@ceng.metu.edu.tr}

\author[hacettepeaddress]{Pinar Duygulu}
\ead{pinar@cs.hacettepe.edu.tr}

\address[hacettepeaddress]{Hacettepe University, Graduate School of Science and Engineering, 06800 Ankara, Turkey}
\address[odtuaddress]{Department of Computer Engineering, Middle East Technical University, 06800 Ankara, Turkey}

\begin{abstract}
Data shift robustness has been primarily investigated from a fully supervised perspective, and robustness of zero-shot learning (ZSL) models have been largely neglected. In this paper, we present novel analyses on the robustness of discriminative ZSL to image corruptions. We subject several ZSL models to a large set of common corruptions and defenses. In order to realize the corruption analysis, we curate and release the first ZSL corruption robustness datasets SUN-C, CUB-C and AWA2-C. We analyse our results by taking into account the dataset characteristics, class imbalance, class transitions between seen and unseen classes and the discrepancies between ZSL and GZSL performances. Our results show that discriminative ZSL suffers from corruptions and this trend is further exacerbated by the severe class imbalance and model weakness inherent in ZSL methods. We then combine our findings with those based on adversarial attacks in ZSL, and highlight the different effects of corruptions and adversarial examples, such as the \textit{pseudo-robustness} effect present under adversarial attacks. We also obtain new strong baselines for both models with the defense methods. Finally, our experiments show that although existing methods to improve robustness somewhat work for ZSL models, they do not produce a tangible effect.

\end{abstract}


\end{frontmatter}

\section{Introduction}
\label{sec:intro}

\noindent Adversarial machine learning, with numerous attack~\cite{xiao2018spatially, su2019one, papernot2016limitations, bhattad2020unrestricted, papernot2017practical, kurakin2016adversarial} and defense~\cite{papernot2016distillation, szegedy2013intriguing, hazan2016perturbations, ross2018improving, lu2017safetynet, naseer2020self, shaham2018defending, guo2018countering} techniques, brought a new perspective to robust generalization ~\cite{xie2020explainable}. Despite the  discussions stemming from adversarial ML, adversarial examples are \textit{worst-case} scenarios and often require \textit{bad intent} to materialize. In addition to adversaries, there are effects which may affect images and the accuracy of a model. A segment of these effects, called \textit{corruptions}, can occur more frequently and naturally, are not \textit{worst-case} scenarios and are not necessarily imperceptible. \cite{hendrycks2019benchmarking} defines a variety of corruptions, and shows that they can invalidate state-of-the-art models in fully-supervised settings.

A significant part of robustness literature in ML has focused on fully supervised models. Zero-Shot Learning (ZSL) and  Generalized Zero-Shot Learning (GZSL) \cite{twostage2009zsl, gbu2018pami} differ from fully supervised settings; in ZSL, the aim is to learn from a set of classes such that the model performs well on classes unseen during the training. GZSL extends this such that the model performs well on both seen and unseen classes. Since ZSL solutions depend on the knowledge transfer between seen and unseen classes, it is already a difficult problem. The introduction of corruptions further exacerbates the issue, increasing the need for a better understanding of ZSL/GZSL robustness.

\begin{figure}[!t]
\begin{center}
      \includegraphics[width=1.0\textwidth]{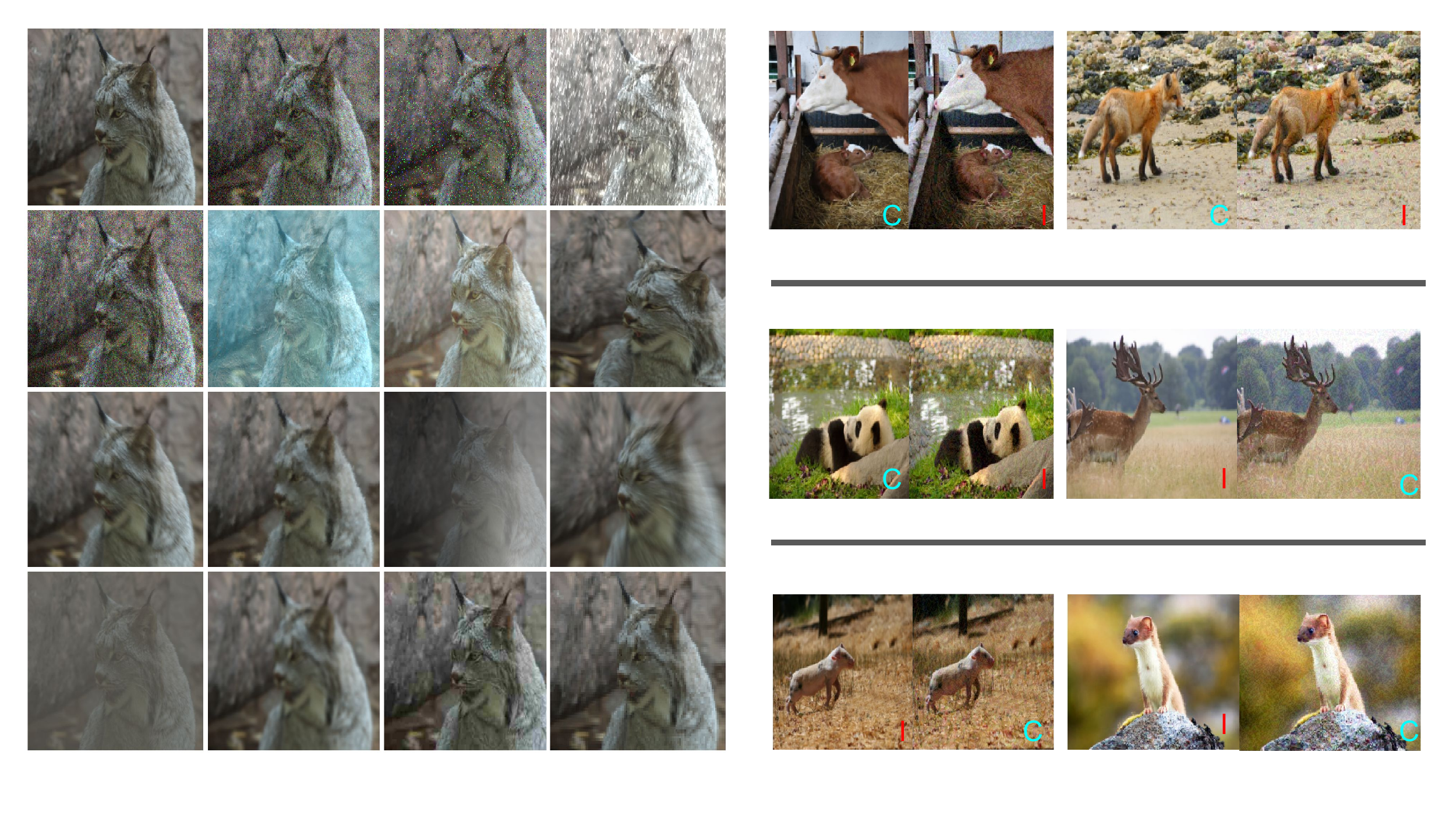}
        \vspace{-16mm}
    \caption{On the left, images from our proposed ZSL corruption datasets (top left, 16 images) are shown. On the right, six image pairs show original and adversarially attacked images, in left and right for each pair, respectively \cite{yucel2020deep}. From top to bottom and left to right, first three pairs show the case of attacks inducing misclassification, whereas the last three show cases where attacks \textit{correct} a misprediction. We empirically show that such \textit{unintuitive} results (\ie attacks correcting a misprediction), which are shown to occur for adversarial attacks and lead to \textit{pseudo-robustness effect} \cite{yucel2020deep}, do not occur for corruptions, and are related to class imbalance and model strength. \textit{I} and \textit{C} indicate incorrect and correct predictions.}
    \label{fig:introduction}
    \vspace{-8mm}
\end{center}
\end{figure}

In this paper, we extend our previous work on adversarial attacks \cite{yucel2020deep} and present rigorous analyses on the robustness of discriminative ZSL models from a corruption perspective.  We leverage the well-established, discriminatively trained label embedding model \cite{akata2013label, weston2010large} and attribute attention model (LFGAA) \cite{liu2019attribute}, and subject them to corruptions and defenses. We also curate and publicly release the first corruption benchmarks for ZSL/GZSL, called CUB-C, SUN-C and AWA2-C, with example cases shown in Figure \ref{fig:introduction}.
Jointly based on our new observations regarding natural corruptions 
and those from \cite{yucel2020deep} regarding adversarial attacks, we conclude our study with important findings and discussions on dataset characteristics, class boundary transitions, severe class imbalance, discrepancies between ZSL and GZSL performances. To sum up, our contributions are as follows.
\begin{itemize}
\item We present a large set of experiments focusing on the robustness of discriminative ZSL models from a corruption perspective. To the best of our knowledge, this is the first study to establish such a benchmark.
\item We curate and release the first benchmark datasets for corruption analysis in ZSL; CUB-C, AWA2-C and SUN-C. The code, datasets and our supplementary material will be made available at \url{https://github.com/MKYucel/zero_shot_corruption_benchmarks}.
\item Our results show discriminative ZSL models are not robust, and we hypothesize the reasons to be severe class imbalance and model weakness. Combined with the results of \cite{yucel2020deep}, we show that the \textit{pseudo-robustness effect}, where absolute metrics may not always reflect the robustness behaviour of a model, is present for adversarial attacks and not for corruptions. This \textit{pseudo-robustness effect} is visualized with examples in Figure \ref{fig:introduction}.
\item We show several defense methods improve the clean accuracy, setting new baselines for both label-attribute embedding and attribute attention models.
\item We show in detail that unseen and seen classes are affected disproportionately by corruptions. We also show zero-shot and generalized zero-shot accuracies are affected differently.
\end{itemize}

Our paper is structured as follows. In Section \ref{related_work}, we review the literature on ZSL, corruption robustness and robust generalization. In Section \ref{methodology}, we motivate our study by presenting our model selection, dataset creation and methods we use to create our benchmark. In Section \ref{experiments}, we present our experimental results and analyses. We perform a comparison of model robustness under adversaries and corruptions in Section \ref{final_words}. We present our final remarks in Section \ref{conclude}.

\section{Related Work} \label{related_work}

\myparagraphwithspace{\textbf{Robust generalization.}} As previously mentioned, adversarial examples often require malicious intent and have arguably low probability of occurrence. Therefore, for a stronger analysis of the robust generalization issue, complementary robustness venues are essential. It has been shown that ImageNet-trained CNNs are biased towards texture, and this can be partially alleviated with a new benchmark Stylized-ImageNet, formed of images with conflicting textures and shapes \cite{geirhos2018imagenet}. Another study released ImageNet-A, which is a curated collection of significantly hard examples of ImageNet classes \cite{hendrycks2019natural}. The degrading effects of distribution shifts have been analysed with various benchmarks, such as ImageNet-R, SVSF and DeepFashion Remixed \cite{hendrycks2020many}.

\myparagraphwithspace{\textbf{Corruption robustness.}} Corruption and perturbation robustness for a range of ML models have been discussed in  \cite{hendrycks2019benchmarking, mu2019mnist}, where ImageNet-C, ImageNet-P, CIFAR-C and MNIST-C benchmarks have been proposed. The corruption robustness benchmarks simulate common image corruptions, such as noise, weather, blur and digital degradations on various severities. In these studies, it has been shown that nearly all state-of-the-art models are invalidated under these corruptions. Several studies showed that data augmentation techniques can help increase robustness \cite{hendrycks2020many,hendrycks2019augmix,calian2021defending}. It is also shown that adversarial training \cite{kireev2021effectiveness}, self-supervised learning \cite{hendrycks2019using}, arbitrary style-transfer \cite{lin2020can}, adversarial noise-training \cite{rusak2020simple} and rectified batch normalization \cite{benz2021revisiting} can improve corruption robustness. Despite the recent advances, corruption robustness is an active field with potential for improvement.

\myparagraphwithspace{\textbf{Zero-shot learning.}} ZSL aims to facilitate learning under extreme data imbalance, where the model is trained on a subset of classes and is required to perform accurately across all classes, even for the ones it has not seen during training. ZSL models achieve this by exploiting auxiliary information, commonly in the form of attributes, to transfer knowledge between seen and unseen classes. Starting with two-stage approaches where first attributes of an image are predicted and these predictions are leveraged to find the class with the most similar attributes \cite{dap2013zsl, twostage2009zsl}, numerous studies formulated one-stage methods where linear \cite{akata2013label, devise2013nips, Akata_2015_CVPR, Li_2018_CVPR} or non-linear \cite{Xian_2016_CVPR, crossmodal2013nips, Chen_2018_CVPR, semantic2018nips} compatibility functions are utilized to map from visual to semantic space. Alternative approaches include mapping from semantic to visual space  \cite{Wang_2018_CVPR, Zhang_2017_CVPR}, embedding visual and semantic information into a shared space \cite{Zhang_2015_ICCV, Changpinyo_2016_CVPR}, transductive methods \cite{Ye_2017_CVPR, trans2017icmr} and discriminative approaches \cite{jiang2019transferable, zhang2018triple}. In recent years, with the increasing success of generative modeling, several methods \cite{sariyildiz2019gradient, bucher2017generating, kumar2018generalized, felix2018multi} produced samples for unseen classes using class embeddings and slowly reduced ZSL to a supervised learning problem. For further information on ZSL, readers are referred to \cite{gbu2018pami,  zsl2019survey}.

A recent preprint \cite{Zhang2019ATZSLDZ} formulates an adversarial training to train ZSL models robust to adversaries. Our study, in contrast, focuses on the intrinsics of discriminative ZSL models from a corruption robustness perspective. Combined with our previous work \cite{yucel2020deep}, we present a comprehensive analysis of discriminative ZSL robustness with different experimental settings where we evaluate not only the effect of corruptions, but also draw insights on dataset characteristics, class transitions and ZSL/GZSL discrepancies. We also present and release the first benchmark datasets for ZSL corruption robustness.

\section{Methodology} \label{methodology}

\noindent In this section, we give the details of our ZSL-robustness benchmarks, explain the defense methods and ZSL formulation that we use in our analyses.

\subsection{Benchmarking ZSL corruption robustness}

The goal of {\em corruption robustness} studies is to understand scenarios where naturally occurring image degradations harm the performance of a model. Therefore, corruption robustness is of great importance for many real-world scenarios. It is critical to have a comprehensive, standardized representation of these effects so that testbeds can be constructed for principled progress in the field.

An important contribution of our work is the curation and release of the corrupted versions of three ZSL datasets, which we name \textit{CUB-C, SUN-C and AWA2-C}.
In defining these benchmarks, we follow the principles of the ImageNet Corruption (ImageNet-C)~\cite{hendrycks2019benchmarking} dataset due to several reasons. First, the corruption types used in generating ImageNet-C sufficiently generalize the possible corruption effects an image may undergo. Second, many ZSL methods use features from ImageNet-pretrained models (\ie ALE) or finetunes ImageNet-pretrained backbones (\ie attribute-attention model). We believe it makes sense to maintain parity with ImageNet-C corruption types to evaluate the robustness of the ZSL methods, as well as the underlying representations.

The proposed \textit{SUN-C, CUB-C and AWA2-C} datasets have four corruption categories (weather, digital, blur and noise) and are formed of 15 corruption types and 5 severity levels. We use the same corruption types used in ImageNet-C \cite{hendrycks2019benchmarking}. We adopt the validation principle of ImageNet-C where models are trained on clean images but validated on four additional corruptions, and then tested on the 15 corruption types.

We adapt the existing corruptions to be suitable to the characteristics of our datasets and make sure they are representative of common detrimental effects (\ie we make sure they are not too weak or too powerful). Finally, due to the significant storage requirements, we create the corruptions on-the-fly in a deterministic manner for reproducibility.

\subsection{Corruption robustness baselines}

In addition to corruptions, we also evaluate several defense methods. We first evaluate three adversarial defense methods against corruptions; label smoothing \cite{labelsmoothadv2020}, total variance minimization (TVM) \cite{guo2018countering} and spatial smoothing \cite{xu2017squeeze} \footnote{Readers are referred to \cite{yucel2020deep} for further details of these techniques.}. Additionally, we test two methods which are effective against image corruptions: AugMix \cite{hendrycks2019augmix} and ANT \cite{rusak2020simple}. AugMix stochastically samples various augmentations, which do not overlap with our corruptions, and enforces similar embeddings via a JS-divergence based consistency loss. ANT uses a noise generator network to learn the most effective noise, which is jointly trained with the classifier.

\subsection{Zero-shot learning model} 
We embrace the well-known label-embedding formulation (ALE) \cite{akata2013label}, which has been shown to be a stable and competitive method even in modern benchmarks \cite{gbu2018pami}, as it is one of the first studies that proposed direct mapping, rather than the previous two-stage approaches. The attribute-label embedding model is formulated as 
\begin{equation}
F(x,y;W) = \theta (x)^{T}W\phi (y) 
\end{equation}
where $\theta(x)$ is the visual embedding and $\phi(y)$ is the class embedding. These two modalities are associated \textcolor{black}{through the compatibility function} $F(.)$, which is parameterized by the learnable weights \textit{W}. We train the model with cross-entropy loss as in \cite{sumbul2019multisource}. We also experiment with the attribute attention model LFGAA \cite{liu2019attribute}, which is a recent and a more accurate model. We choose this method because it performs a non-linear mapping (via backbone finetuning) to both semantic and latent feature spaces, and has inductive and transductive variants. We believe the results of ALE and LFGAA will be representative of several families of discriminative ZSL approaches. 

We acknowledge that especially in GZSL setting, 
most recent state-of-the-art works rely on generative approaches, where unseen class samples are generated using class embedding conditional generative models~\cite{sariyildiz2019gradient, bucher2017generating, kumar2018generalized, felix2018multi}. Such approaches require complex pipelines where synthetic samples (in feature space) are generated and models are expected to learn from synthetic samples and perform well on real samples. Although such methods present interesting venues
for robustness analyses, their complex pipelines complicate the study of robustness. Therefore, we keep the generative approaches outside the scope of our work and focus on discriminatively trained ZSL models.

\section{Experiments} \label{experiments}

\noindent In this section, we give the experimental setup and implementation details, and then present our results and analyses.

\subsection{Datasets and evaluation metrics}

We perform our evaluation on three ZSL/GZSL datasets; Caltech-UCSD-Birds 200-2011 (CUB) \cite{cubdataset}, Animals with Attributes 2 (AWA2) \cite{gbu2018pami} and SUN \cite{patterson2014sun}. CUB is a medium-sized dataset with 312 attributes, 200 classes and 11788 images. As classes are similar to each other appearance-wise and each class has few samples, it represents a challenging case. SUN is another medium-sized dataset that has 102 attributes, 717 classes and 14340 images. Due to having quite different classes and even lower per-class sample count than CUB, SUN is also a challenging dataset. AWA2, on the other hand, is a larger dataset with 85 attributes, 50 classes and 37322 images. Although it has a good per-class sample count, this leads to a more pronounced class imbalance between seen and unseen classes, which makes it a challenging case for ZSL/GZSL settings. We use the splits proposed in \cite{gbu2018pami}. We use the corrupted versions \textit{SUN-C, CUB-C and AWA2-C} and evaluate our models using 15 corruptions with 5 severity levels. 

In the corruption robustness literature, \textit{Mean Corruption Error} (MCE) is the common metric; MCE calculates the errors for each corruption type using all severity levels and weighs them using AlexNet errors, and then takes the mean of each corruption category to produce the final MCE score \cite{hendrycks2019benchmarking}. Due to the sheer volume of experiments in our case, providing MCE values is not feasible \footnote{We provide MCE values in our supplementary material.}. Instead, we first calculate the corrupted accuracy values for each corruption type and severity level, find the reduction in accuracy compared to the original accuracy, and then calculate the ratio of reduction (i.e. as a percentage of the original non-corrupted results) for each corruption category \footnote{Our results are reminiscent of Relative MCE \cite{hendrycks2019benchmarking}, which is another metric for corruption robustness.}, as shown in Figures \ref{fig:corruption_graph} and \ref{fig:corruption_graph_lfgaa}.  For defense methods, we do the same with a key difference; instead of finding the reduction in accuracy compared to the original accuracy, we find the reductions in accuracy (produced by the defense method) compared to the corrupted accuracy values (for each severity and type) and then calculate the ratio of reduction (i.e. as a percentage of the original non-corrupted results) for each corruption category, as shown in Figures \ref{fig:combined_1}, \ref{fig:combined_2}, \ref{fig:transductive_defenses} and \ref{fig:inductive_defenses} \footnote{We note that Figures \ref{fig:combined_1} and \ref{fig:combined_2} can be compared to each other but not to Figure \ref{fig:corruption_graph}.}. Finally, we believe our results can be the \textit{ZSL-version} of AlexNet error-weighting scheme for ZSL corruption robustness, due to the prevalence of ALE model.

\subsection{Implementation details} \label{implementation_details}

We merge ResNet-101 \cite{he2016deep} feature extractor with ALE. We train the ALE model, keeping the feature extractor fixed.  For LFGAA, we retrain all the models for all datasets for both inductive and transductive variants. We use ResNet backbone for CUB and SUN, and VGG for AWA2. We indicate the reproduced values as \textit{LFGAA+SA} and \textit{LFGAA+Hybrid} for LFGAA, and \textit{original} for ALE, although there are slight variations coompared to original results \cite{liu2019attribute,gbu2018pami}. 
We use PyTorch \cite{paszke2019pytorch} for our experiments.

For label smoothing, we assign 0.9 to the ground-truth. For spatial smoothing, we use 3x3 windows. The maximum iteration count is chosen as 3 for TVM. We apply the same corruption and defense parameters to all datasets for a fair comparison. We first resize and then corrupt the images to make sure a comparable effect is achieved for all images.  We use the original implementations for AugMix (with JS divergence) and ANT, and tune the results on the validation corruptions.

\subsection{Corruption robustness experiments - ALE} \label{corruption_results}
\begin{figure}[!t]
\begin{center}
\renewcommand{\arraystretch}{1} 
      \includegraphics[width=1.\textwidth]{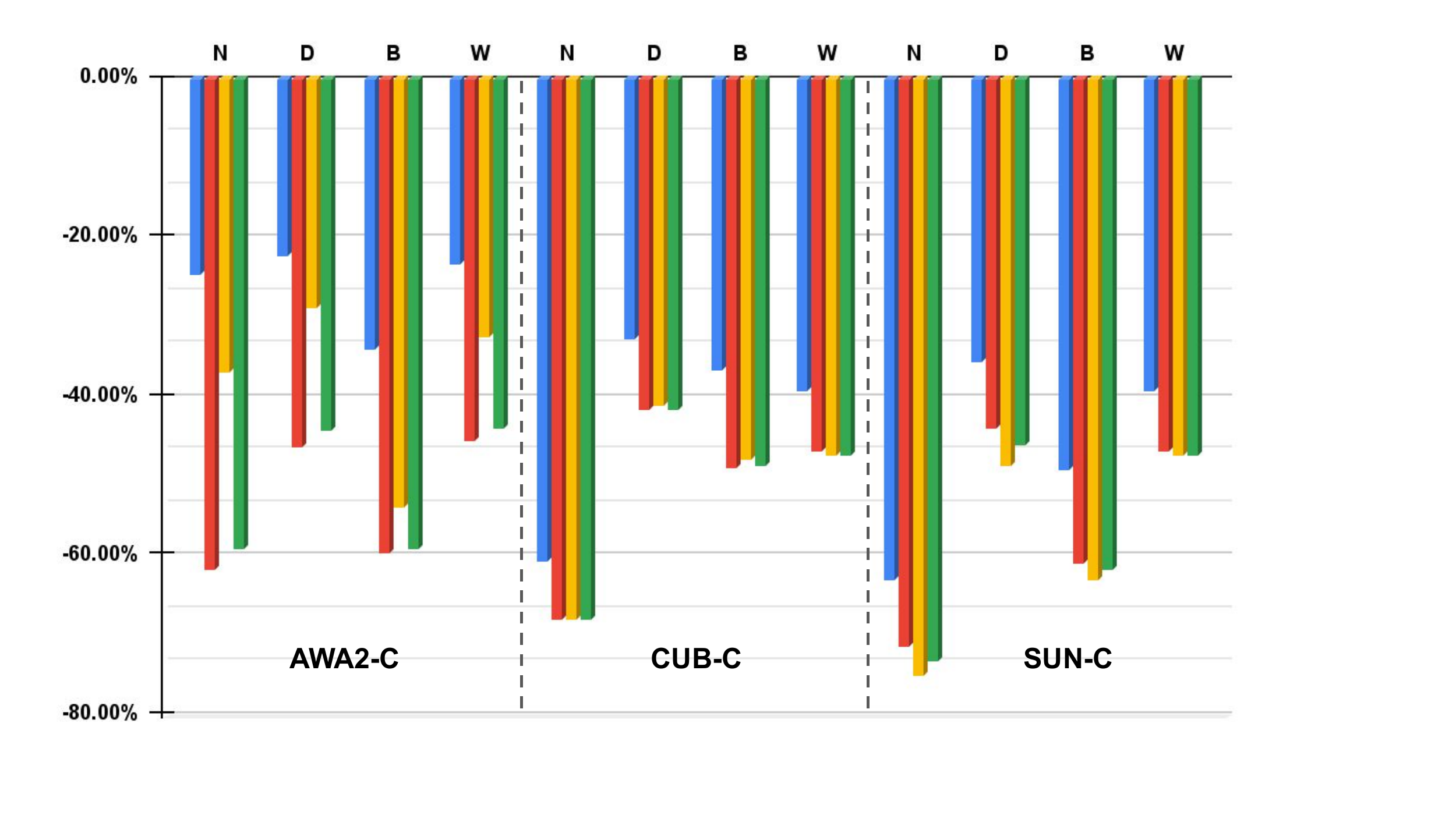}
        \vspace{-15mm}
    \caption{Corruption induced category-based average reductions, as a percentage of the original (non-corrupted) results. \textit{N, D, B} and \textit{W} are \textit{noise}, \textit{digital}, \textit{blur} and \textit{weather} categories, respectively. Blue, red, orange and green bars indicate ZSL top-1, unseen, seen and harmonic scores, respectively.}
    \label{fig:corruption_graph}
    \vspace{-5mm}
\end{center}
\end{figure}

\myparagraphwithspace{\textbf{Corruptions.}} 
The impact of corruptions on ALE is shown in Figure \ref{fig:corruption_graph}. In the \textit{ZSL} setting, we see across the board reductions of accuracy values. On AWA2, highest ZSL reduction occurs in blur category. On CUB, noise category introduces a dramatic reduction in accuracy values up to 60\%. SUN experiences the highest reduction in accuracy (60\%) when exposed to noise. For all datasets, digital corruptions introduce the least amount of impact. On individual corruption types, we observe brightness to be the weakest one. Gaussian noise, shot noise and contrast corruptions introduce the highest accuracy reductions on CUB, SUN and AWA2 datasets, respectively.

In the \textit{GZSL} setting, corruptions introduce significant reductions across all datasets. On AWA2, noise and blur introduce the same degradation in harmonic scores (60\%), although effects on unseen and seen classes for noise and blur vary. On CUB and SUN, noise introduces the highest reduction in harmonic scores (up to 75\% on SUN). On AWA2, unseen classes suffer more than seen classes. On CUB, seen and unseen classes are affected similarly, whereas on SUN seen classes are affected slightly more. On individual corruption types, brightness is still the weakest. Contrast, impulse and Gaussian noise introduce the biggest impact on accuracy on AWA2, CUB and SUN, respectively.

\myparagraphwithspace{\textbf{Defenses.}} The spatial smoothing results are shown in Figure \ref{fig:combined_1}.  In the \textit{ZSL} setting, spatial smoothing degrades the results across all datasets. On AWA2, noise is affected the worst (-7.5\%) whereas digital is the least affected (-3.5\%). On CUB and SUN, weather and noise are the worst and least affected, respectively. Spatial smoothing only works on SUN dataset for noise (+1.5\%). On individual corruption types, spatial smoothing only works well against impulse noise for all datasets. Spatial smoothing fails to work for \textit{GZSL} as well. Across all datasets, weather and noise are the worst and the least affected, respectively. The only time spatial smoothing works is for noise on SUN dataset; it improves unseen classes and harmonic scores (0.5\%). Similar to ZSL, spatial smoothing only works against impulse noise for all datasets in GZSL.

The \textit{total variance minimization} results are shown in Figure \ref{fig:combined_1}. In the \textit{ZSL} setting, TVM fails to introduce improvements, but does a slightly better job than spatial smoothing. Across all datasets, we see the noise results either improved (\ie +3.4\% on SUN) or affected minimally whereas weather category is affected the worst. We see recoveries for noise on CUB and SUN, and also blur shows improvements on SUN. On individual corruption types, we see all noise types experiencing improvements in all datasets. On AWA2, some blur types are recovered as well, especially in high severity levels. On CUB, all noise types and glass blur are improved. On SUN, across the board improvements are observed for all noise and blur corruptions. In the \textit{GZSL} setting, the results are worse, but better than spatial smoothing. Across all datasets, weather is the worst affected one (\ie  -10\% on AWA2 h-score). Noise and blur are the best for AWA2/CUN and SUN, respectively. We see improvements only in noise and blur for CUB and SUN, respectively. We see unseen class accuracies spearheading the recoveries when there is an improvement (\ie +2.5\% in unseen on SUN for blur). On individual corruption types, we see zoom blur to be the only improved corruption on AWA2. All noise types and glass blur experiences recoveries on CUB. On SUN, all blur types are improved. Unseen classes still undergo less degradation than seen classes, especially on SUN and AWA2 experiments.

For label smoothing, AugMix and ANT, we retrain the models and report the results in Table \ref{corruption_defense_table}. Label smoothing has a slight degrading effect, except on AWA2 GZSL where it improves the unseen (+1 point) and harmonic scores (+1.1 point). AugMix performs worse than label smoothing in ZSL but manages to improve ZSL top-1 on SUN dataset by 1 point. In GZSL, AugMix improves both seen, unseen and harmonic scores, except on CUB where seen class accuracy is reduced by 4.5 points. ANT has a similar trend with AugMix, it reduces ZSL scores. In GZSL, it improves unseen classes for all datasets and improves harmonic scores on AWA2 by around 6 points and SUN by 3.3 points.

\begingroup
\setlength{\tabcolsep}{8pt} 
\renewcommand{\arraystretch}{1} 
\begin{table}[!t]
\resizebox{\textwidth}{!}{%
\begin{tabular}{ccccccccccccc}
\multicolumn{1}{l}{} & \multicolumn{3}{c|}{\textbf{Zero Shot}}  & \multicolumn{9}{c}{\textbf{Generalized Zero Shot}}                                                        \\

\multicolumn{1}{l}{} & C    & S    & \multicolumn{1}{c|}{A}    & \multicolumn{3}{c|}{C}                  & \multicolumn{3}{c|}{S}                  & \multicolumn{3}{c}{A} \\ \hline
               & \multicolumn{3}{c|}{Top-1}              & u    & s    & \multicolumn{1}{c|}{h}    & u    & s    & \multicolumn{1}{c|}{h}    & u     & s     & h     \\ \hline
Original             & \textbf{54.5} & 57.4 & \multicolumn{1}{c|}{\textbf{62.0}} & 25.6 & \textbf{64.6} & \multicolumn{1}{c|}{36.7} & 20.5 & 32.3 & \multicolumn{1}{c|}{25.1} & 15.3  & 78.8  & 25.7  \\ \hline

LbS         & 52.2 & 55.2 & \multicolumn{1}{c|}{60.6} & 22.7 & 56.2 & \multicolumn{1}{c|}{32.4} & 18.4 & 31.6 & \multicolumn{1}{c|}{23.3} & 16.3  & 74.2  & 26.8  \\

AugMix       & 51.6  & \textbf{58.4} & \multicolumn{1}{c|}{54.9} & \textbf{27.2} &  60.1 & \multicolumn{1}{c|}{\textbf{37.5}} & \textbf{23.6} & 35.7 & \multicolumn{1}{c|}{\textbf{28.4}} & 16.1  & 83.5  & 27.0  \\

ANT      & 48.9 & 57.4 & \multicolumn{1}{c|}{55.1} & 26.1 & 60.6 & \multicolumn{1}{c|}{36.4} & \textbf{23.6} & \textbf{35.8} & \multicolumn{1}{c|}{\textbf{28.4}} & \textbf{19.3}  & \textbf{85.4}  & \textbf{31.6} 
\end{tabular}
}
\caption{Results on CUB (C), SUN (S) and AWA2 (A) of ALE models trained with \textit{label smoothing}, \textit{AugMix} and \textit{ANT} approaches.}
\label{corruption_defense_table}
\end{table}
\endgroup

The label smoothing results are shown in Figure \ref{fig:combined_1}. In \textit{ZSL},  we see minimal improvements in isolated cases (\ie weather category on CUB,  +0.3\%). On AWA2, all categories are affected similarly but weather is the worst affected (-5\%). On CUB, weather results are improved slightly and noise is the worst affected one. On SUN, all categories reduce the accuracy but with a slight margin. In \textit{GZSL}, we see improvements, especially on AWA2. Unseen and harmonic scores are recovered visibly (up to 12\%), possibly owing to the increase in unseen values shown in Table \ref{corruption_defense_table}, but seen accuracies get worse. Similar to AWA2, weather enjoys the best results on CUB, though it goes through accuracy reduction. On SUN, we see minor improvements in seen accuracies for digital (0.2\%) and weather (1.5\%), whereas unseen accuracies go through severe degradation. On individual corruption types, the trends follow the category-wise results.

\begin{figure}[!t]
\begin{center}
\renewcommand{\arraystretch}{1} 
      \includegraphics[width=1.\textwidth]{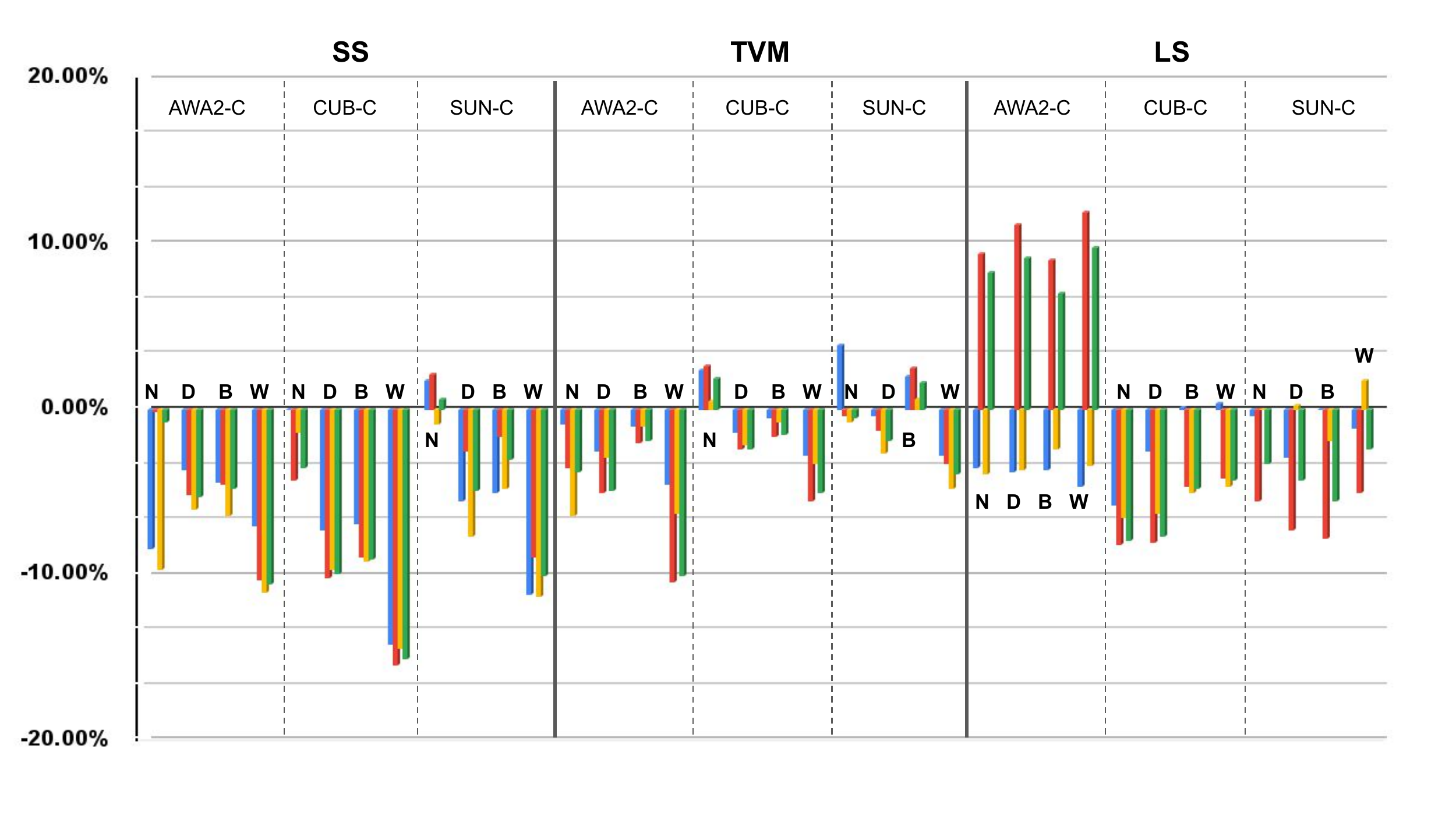}
        \vspace{-15mm}
    \caption{The (relative) effect of \textit{spatial smoothing (SS), total-variance minimization (TVM) and label smoothing(LS)} defenses over corruptions, provided as category-based average reductions as a percentage of the original (non-corrupted) results. \textit{N, D, B} and \textit{W} are \textit{noise}, \textit{digital}, \textit{blur} and \textit{weather} categories, and blue, red, orange and green bars indicate ZSL top-1, unseen, seen and harmonic scores, respectively. Positive ratios indicate that the defenses improves over the corruptions, whereas the negative ratios indicate that the defenses produce even lower results over the corruptions.}
    \label{fig:combined_1}
    \vspace{-5mm}
\end{center}
\end{figure}

\begin{figure}[!t]
\begin{center}
\renewcommand{\arraystretch}{1} 
      \includegraphics[width=1.\textwidth]{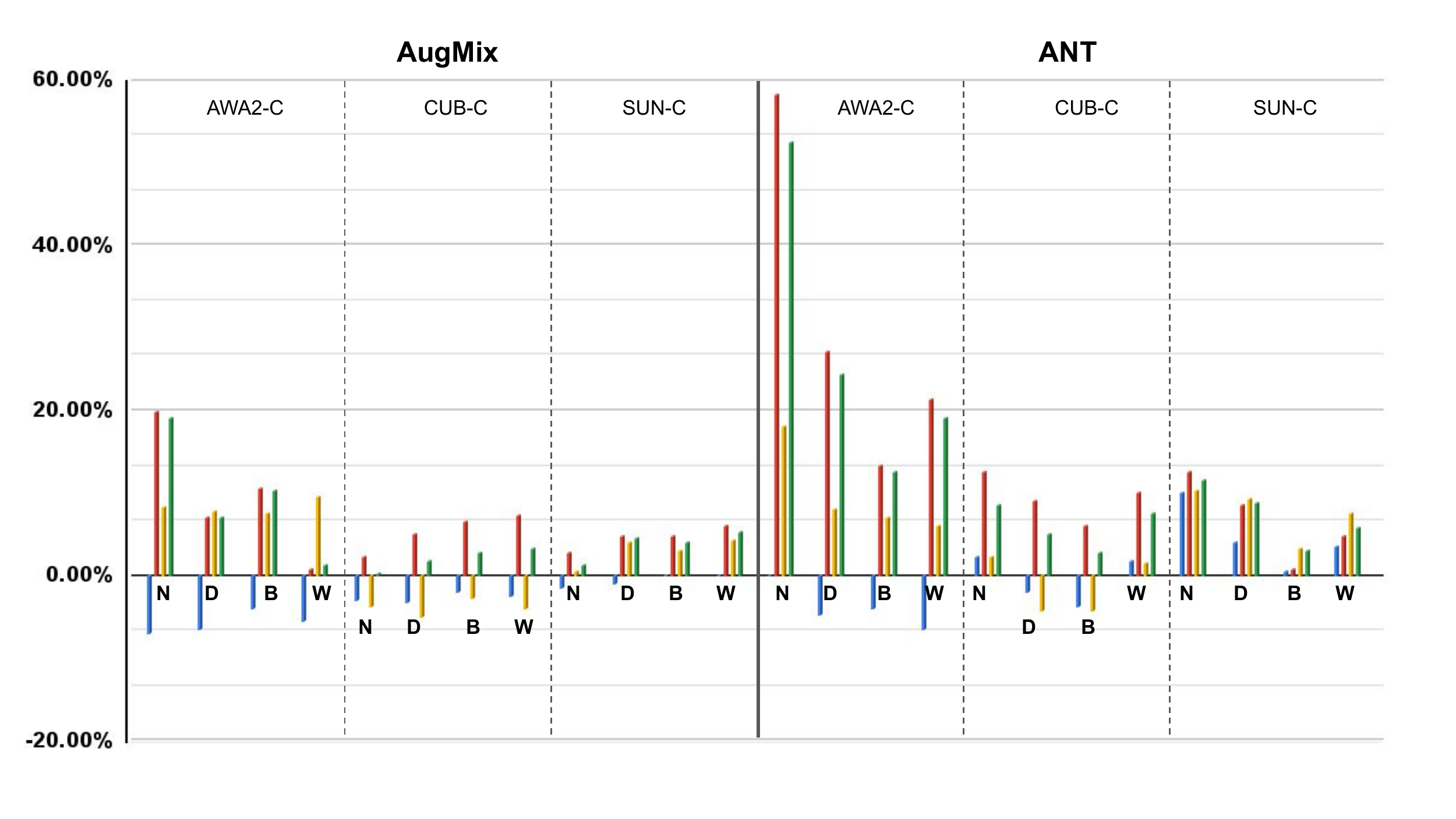}
        \vspace{-15mm}
    \caption{The (relative) effect of \textit{AugMix and ANT} defenses over corruptions, provided as category-based average reductions as a percentage of the original (non-corrupted) results. Refer to Figure \ref{fig:combined_1} for legend details.}    
    \label{fig:combined_2}
    \vspace{-5mm}
\end{center}
\end{figure}

The AugMix results are shown in Figure \ref{fig:combined_2}. In \textit{ZSL}, AugMix fails to perform. Noise categories is the worst affected one, except digital which is slightly worse than noise on CUB. Blur category enjoys the smallest degradation across all datasets. Despite not seeing improvements in ZSL accuracies, the detrimental effects are weaker than the previous defenses; on SUN, blur and digital barely introduce further degradation (-0.2\% for both). In the \textit{GZSL} setting, we see significant improvements across all datasets.  Except seen classes on CUB, where reductions up to 5\% are observed, both seen and unseen classes are recovered across all datasets. Except some isolated cases (digital and weather on AWA2), unseen class recoveries are significantly better than seen classes. Compared to label smoothing, AWA2 recoveries are slightly worse in average but AugMix is the first method which provides consistent recoveries. Noise and weather categories experience the highest recovery rates for AWA2 (+18\% in h-score) and SUN/CUB (up to +5\% in h-score), respectively. On individual corruption types, the trends are similar to category-level results.

The results of ANT are shown in Figure \ref{fig:combined_2}. In \textit{ZSL}, ANT performs the best. The results show consistent recoveries on SUN, partial recoveries on CUB and slight degradations on AWA2.  On AWA2, noise performs the best whereas weather is the worst (-7\%). Blur goes through degradation (-4\%) and slight improvement (+1.2\%) on CUB and SUN, respectively. Noise performs the best and introduces visible improvements in both CUB (+3\%) and SUN (+10\%). In the \textit{GZSL} setting, we see significant and consistent recoveries, which makes ANT the best performing method. Across all datasets and noise categories, unseen and harmonic accuracies are improved. Seen classes, except digital and blur on CUB (-5\%), experience recoveries as well. Unseen/seen discrepancy is visible and generally, unseen classes enjoy better returns with ANT.

\myparagraphwithspace{\textbf{Summary.}} We see that the digital corruptions to be the weakest. Noise introduces the most dramatic reductions, where reductions of 60\% on AWA2 h-score and 73\% on SUN h-score are observed. Brightness corruption is the weakest  across the board, whereas different noise types and surprisingly contrast (a digital corruption) are the most effective corruption types. In defenses, ANT is the most effective defense. Our results show spatial smoothing to be most ineffective defense.

\subsection{Corruption robustness experiments - LFGAA}

\begin{figure}[!t]
\begin{center}
\renewcommand{\arraystretch}{1} 
      \includegraphics[width=1.\textwidth]{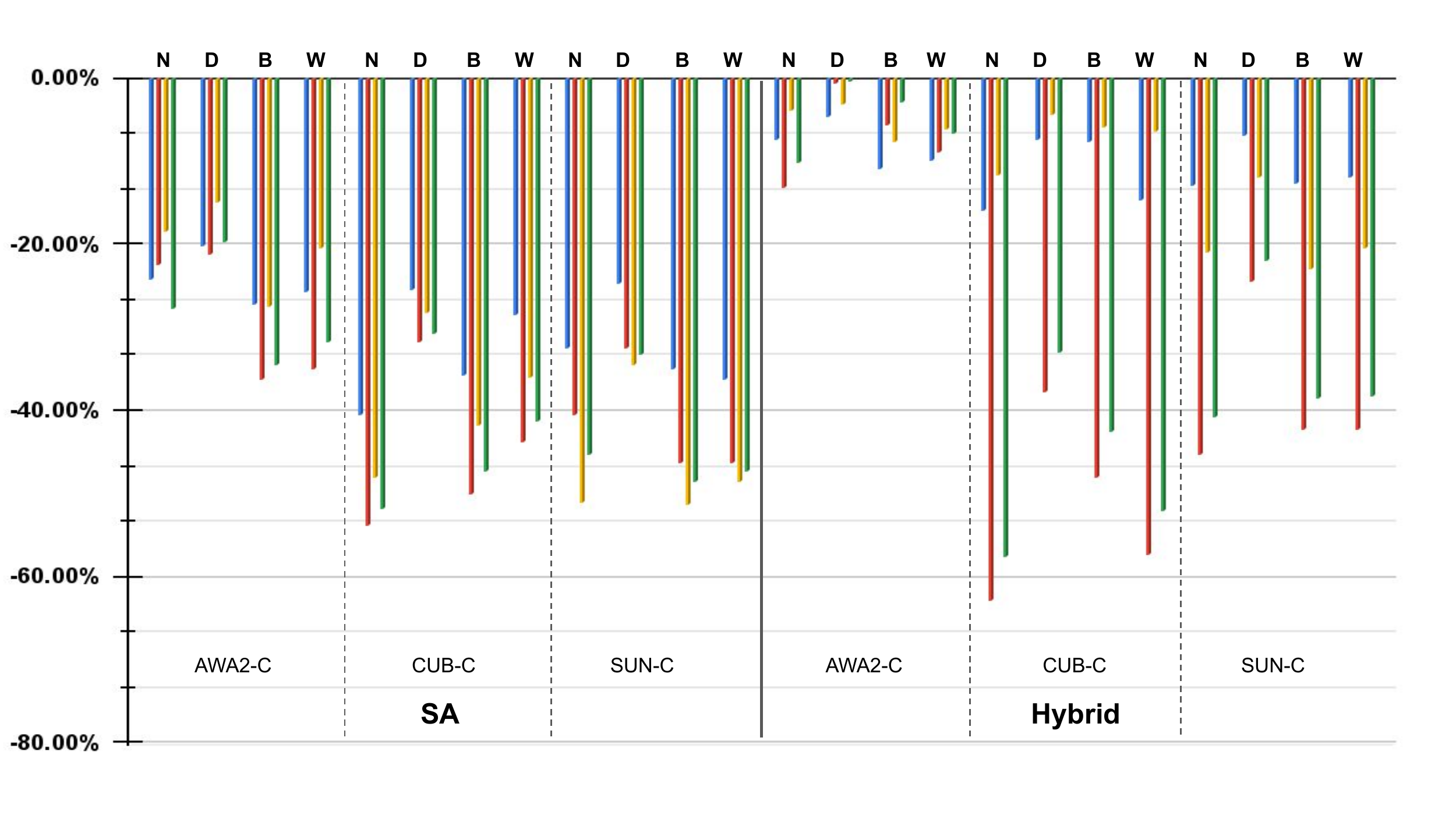}
        \vspace{-15mm}
    \caption{Corruption induced category-based average reductions, as a percentage of the original (non-corrupted) results for LFGAA \cite{liu2019attribute}. Refer to Figure \ref{fig:corruption_graph} for legend details.  \textit{SA} indicates self-adaptation, whereas \textit{Hybrid} is the inductive variant.}
    \label{fig:corruption_graph_lfgaa}
    \vspace{-5mm}
\end{center}
\end{figure}

\myparagraphwithspace{\textbf{Corruptions.}} The impact of corruptions on LFGAA is shown in Figure \ref{fig:corruption_graph_lfgaa}. In the \textit{ZSL} setting, the transductive variant (\textit{LFGAA+SA}) shows across the board reductions, where the reductions are least prominent in AWA2. The inductive variant (\textit{LFGAA+Hybrid}) proves to be more robust than \textit{SA}; especially in AWA2 the effect of the corruptions are considerably smaller (-5\% in digital). In the \textit{GZSL} setting, both variants show reductions across the board. Except SUN in \textit{LFGAA+SA}, nearly all other results show that unseen classes go through a more severe degradation. \textit{LFGAA+Hybrid} shows slightly less degradations than \textit{LFGAA+SA} in the \textit{GZSL} setting, which is quite visible for AWA2 results. Especially in \textit{LFGAA+Hybrid}, unseen classes are affected quite severely.

\myparagraphwithspace{\textbf{Defenses.}}  We provide the accuracy values of LFGAA trained with AugMix and ANT in Table \ref{lfgaa_corruption_defense_table}, as well as the reproduced original results. We see that the effects on AugMix and ANT are mixed. AugMix improves on CUB ZSL (+2.6\%) and AWA2 GZSL (+7.5\%) for \textit{LFGAA+SA} and improves on AWA2 GZSL (+5\%) for \textit{LFGAA+Hybrid}. ANT performs better than AugMix, where it shows improvements on ZSL and GZSL for all datasets except SUN GZSL for \textit{LFGAA+SA}. In \textit{LFGAA+Hybrid}, it shows minimal degradations for ZSL but shows consistent improvements in GZSL.

\begingroup
\setlength{\tabcolsep}{8pt} 
\renewcommand{\arraystretch}{1} 
\begin{table}[!t]
\resizebox{\textwidth}{!}{%
\begin{tabular}{ccccccccccccc}
\multicolumn{1}{l}{} & \multicolumn{3}{c|}{\textbf{Zero Shot}}  & \multicolumn{9}{c}{\textbf{Generalized Zero Shot}}                                                        \\
\multicolumn{1}{l}{} & C    & S    & \multicolumn{1}{c|}{A}    & \multicolumn{3}{c|}{C}                  & \multicolumn{3}{c|}{S}                  & \multicolumn{3}{c}{A} \\ \hline
               & \multicolumn{3}{c|}{Top-1}              & u    & s    & \multicolumn{1}{c|}{h}    & u    & s    & \multicolumn{1}{c|}{h}    & u     & s     & h     \\ \hline
LFGAA+SA            & 78.9 & 58.7 & \multicolumn{1}{c|}{74.9} & 43.4 & \textbf{79.6} & \multicolumn{1}{c|}{56.2} & \textbf{16.0} & \textbf{26.0} & \multicolumn{1}{c|}{\textbf{19.8}} & 33.2  & 83.3  & 47.5  \\ 

 + AugMix       & \textbf{81.5} & 55.0 & \multicolumn{1}{c|}{65.0} & 44.2 &  65.4 & \multicolumn{1}{c|}{52.7} & 14.7 & 20.6 & \multicolumn{1}{c|}{17.2} & 42.0  & 80.1 & 55.1  \\

 + ANT      & 80.6 & \textbf{59.5} & \multicolumn{1}{c|}{\textbf{75.6}} & \textbf{50.0} & 69.0 & \multicolumn{1}{c|}{\textbf{57.9}} & 15.8 & 23.2 & \multicolumn{1}{c|}{18.8} & \textbf{43.3 } & \textbf{89.0}  & \textbf{58.2} \\ \hline

LFGAA+Hybrid             & \textbf{71.6} &\textbf{ 57.9} & \multicolumn{1}{c|}{\textbf{70.4}} & 25.9 & \textbf{81.8} & \multicolumn{1}{c|}{39.4} & 14.0 & \textbf{39.7} & \multicolumn{1}{c|}{20.8} & 15.2  & 88.4  & 25.9  \\

 + AugMix       & 66.1  & 56.3 & \multicolumn{1}{c|}{64.6} & 20.9 &  78.5 & \multicolumn{1}{c|}{33.0} & 13.4 & \textbf{39.7} & \multicolumn{1}{c|}{20.0} & \textbf{18.6}  & 91.3  & \textbf{31.0}  \\

 + ANT      & 70.7 & 56.6 & \multicolumn{1}{c|}{68.8} & \textbf{28.0} & 80.4 & \multicolumn{1}{c|}{\textbf{41.6}} & \textbf{15.4} & 38.8 & \multicolumn{1}{c|}{\textbf{22.1}} & 16.4  & \textbf{92.5}  & 27.8 \\

\end{tabular}
}
\caption{Results of LFGAA \cite{liu2019attribute} models trained with \textit{AugMix} and \textit{ANT} approaches.}
\label{lfgaa_corruption_defense_table}
\end{table}
\endgroup

\begin{figure}[!t]
\begin{center}
\renewcommand{\arraystretch}{1} 
      \includegraphics[width=1.\textwidth]{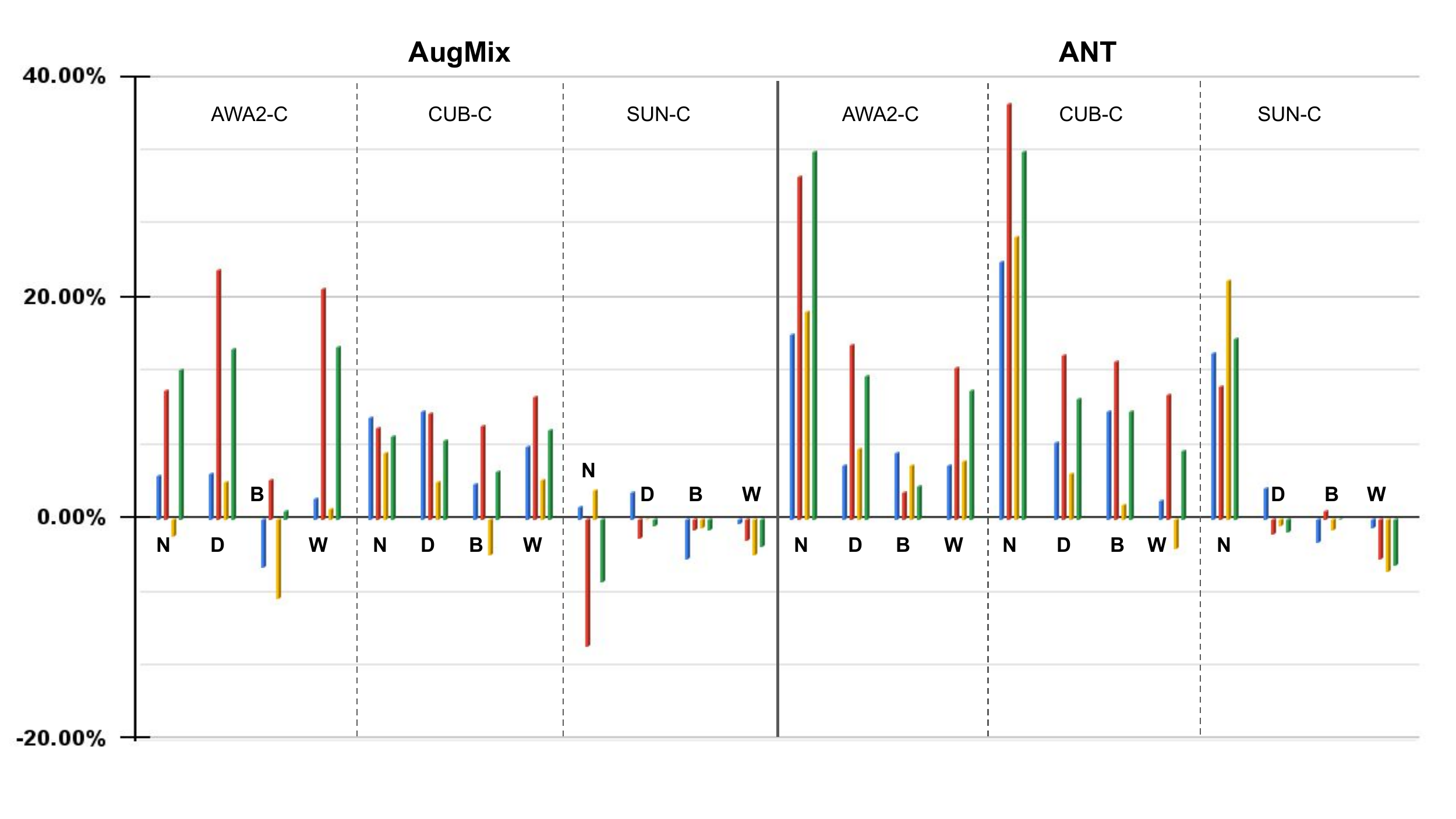}
        \vspace{-15mm}
    \caption{The (relative) effect of \textit{AugMix and ANT} defenses over corruptions, provided as category-based average reductions as a percentage of the original (non-corrupted) results for LFGAA \cite{liu2019attribute} (with self-adaptation). Refer to Figure \ref{fig:combined_1} for legend details.}
    \label{fig:transductive_defenses}
    \vspace{-5mm}
\end{center}
\end{figure}

\begin{figure}[!t]
\begin{center}
\renewcommand{\arraystretch}{1} 
      \includegraphics[width=1.\textwidth]{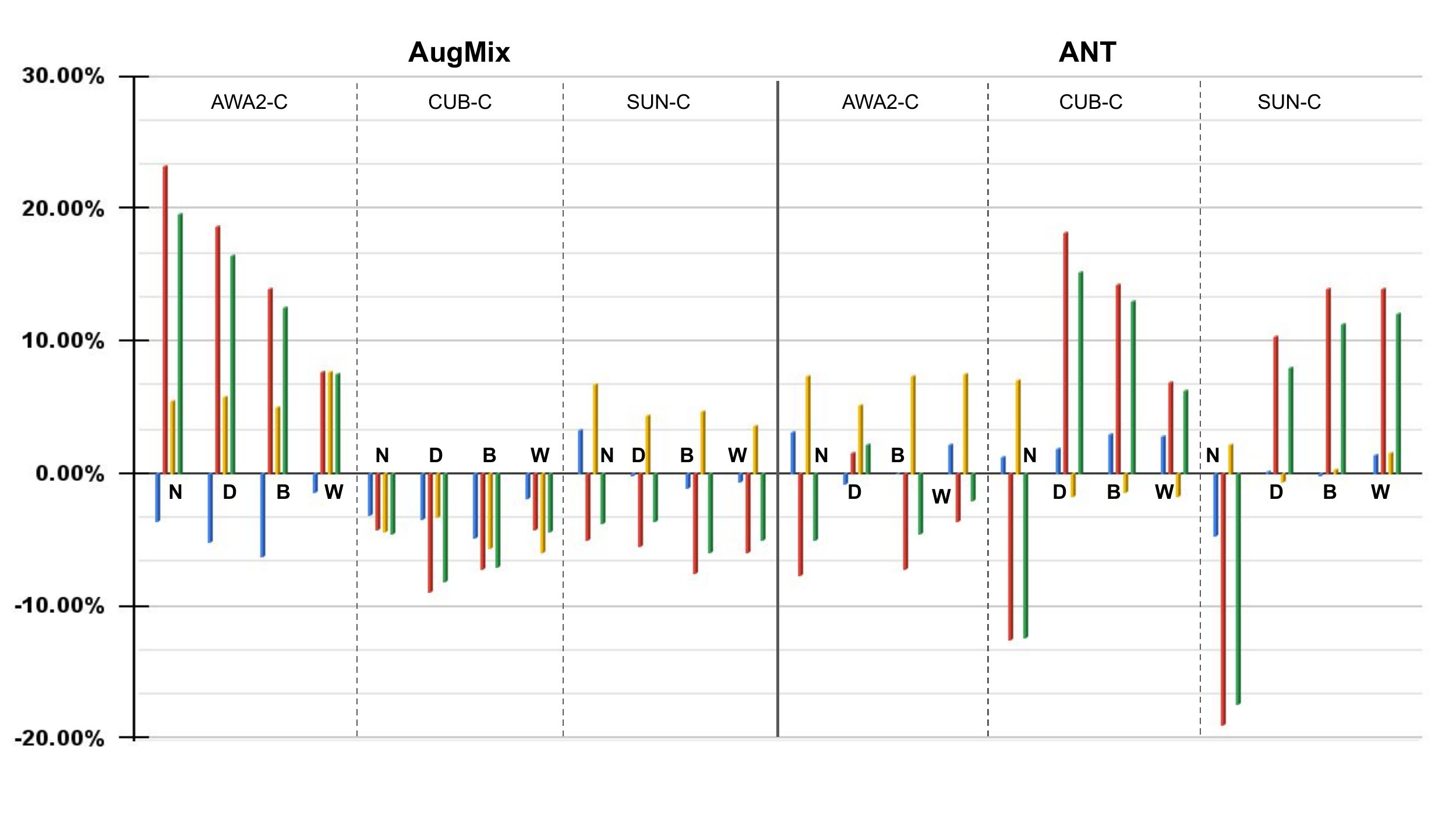}
        \vspace{-15mm}
    \caption{The (relative) effect of \textit{AugMix and ANT} defenses over corruptions, provided as category-based average reductions as a percentage of the original (non-corrupted) results for LFGAA \cite{liu2019attribute} (hybrid). Refer to Figure \ref{fig:combined_1} for legend details.}
    \label{fig:inductive_defenses}
    \vspace{-5mm}
\end{center}
\end{figure}

AugMix, for \textit{LFGAA+SA} (see Figure \ref{fig:transductive_defenses}), introduces improvements in ZSL except some categories, especially on SUN. In the \textit{GZSL} setting, except SUN, it introduces visible improvements. The results also show that AugMix helps recover unseen class accuracies better than seen classes. For \textit{LFGAA+Hybrid} (see Figure \ref{fig:inductive_defenses}), AugMix does a worse job; in ZSL it degrades the results further, except some cases (i.e. noise in SUN). In the \textit{GZSL} setting, the results look better for AWA2 but for other datasets the degradation trend is visible.

ANT, for \textit{LFGAA+SA} (see Figure \ref{fig:transductive_defenses}), introduces better results. In ZSL, except some isolated cases on SUN, consistent improvements are observed. In the \textit{GZSL} setting the trend is the same. The recoveries in unseen classes are visibly better in average. For \textit{LFGAA+Hybrid} (see Figure \ref{fig:inductive_defenses}), ANT fails to recover. In ZSL, it introduces degradations but does a better job than AugMix. In the \textit{GZSL} setting, except AWA2 and some other cases (i.e. noise in CUN and SUN), it actually introduces improvements. In overall, ANT does a better job, but fails to provide a tangible improvement in \textit{LFGAA+Hybrid}.

\subsection{Analysing corruption robustness} \label{corruption_analyses}

Our corruption robustness results show similarities with our observations on adversarial attacks \cite{yucel2020deep}, in terms of unseen/seen and ZSL/GZSL discrepancy, and adverse effect of defenses.  We perform our analyses to see the characteristics of ZSL models under the effect of corruptions. The analyses presented here are primarily for the ALE model, but the findings apply to LFGAA as well.

\begingroup
\setlength{\tabcolsep}{6pt} 
\renewcommand{\arraystretch}{1} 
\begin{table}[]
\resizebox{\textwidth}{!}{%
\begin{tabular}{cclcclccccllcclllll}
\multicolumn{1}{l}{} & \multicolumn{18}{c}{\textbf{Generalized Zero Shot}}                                                                                                                                                                                                                                                                               \\
\multicolumn{1}{l}{} & \multicolumn{6}{c|}{C}                                                                                        & \multicolumn{6}{c|}{S}                                                                                                            & \multicolumn{6}{c}{A}                                                         \\ \hline
Class Type           & \multicolumn{3}{c|}{U}                                & \multicolumn{3}{c|}{S}                                & \multicolumn{3}{c|}{U}                                                    & \multicolumn{3}{c|}{S}                                & \multicolumn{3}{c|}{U}                                & \multicolumn{3}{c}{S} \\ \hline
Transitions          & \multicolumn{1}{l}{CF} & FC & \multicolumn{1}{l|}{FF} & \multicolumn{1}{l}{CF} & FC & \multicolumn{1}{l|}{FF} & \multicolumn{1}{l}{CF} & \multicolumn{1}{l}{FC} & \multicolumn{1}{l|}{FF} & CF                     & FC & \multicolumn{1}{l|}{FF} & \multicolumn{1}{l}{CF} & FC & \multicolumn{1}{l|}{FF} & CF    & FC    & FF    \\ \hline

$Noise$             & 79                     & 4 & \multicolumn{1}{c|}{78} & 73                     & 7 & \multicolumn{1}{c|}{79} & 81                     & 3                     & \multicolumn{1}{c|}{85} & \multicolumn{1}{c}{81} & 3 & \multicolumn{1}{c|}{84} & 75                     & 2 & \multicolumn{1}{l|}{61} & 38    & 18    & 47    \\

$Digital$             & 59                     & 6 & \multicolumn{1}{c|}{68} & 48                     & 12 & \multicolumn{1}{c|}{59} & 60                     & 4                     & \multicolumn{1}{c|}{67} & 58                     & 5 & \multicolumn{1}{c|}{67} & 64                    & 3  & \multicolumn{1}{l|}{50} & 30    & 16    & 39    \\ \hline

$Weather$             & 61                     & 5 & \multicolumn{1}{c|}{70} & 53                     & 10 & \multicolumn{1}{c|}{62} & 63                     & 4                    & \multicolumn{1}{c|}{71} & 61                     & 4 & \multicolumn{1}{c|}{69} & 60                     & 3  & \multicolumn{1}{l|}{51} & 33    & 16    & 42    \\

$Blur$             & 63                     & 5 & \multicolumn{1}{c|}{72} & 55                     & 12 & \multicolumn{1}{c|}{64} & 83                     & 3                      & \multicolumn{1}{c|}{79} & 71                     & 4 & \multicolumn{1}{c|}{78} & 79                     & 3  & \multicolumn{1}{l|}{68} & 54    & 14    & 55    \\ \hline
\end{tabular}%
}
\caption{Categorization of prediction changes induced by each corruption category. U and S columns are results for unseen and seen classes, respectively. CF, FC and FF are \textit{correct-to-false } (as the percentage of all originally correct predictions), \textit{false-to-correct} and \textit{false-to-other-false} (as the percentage of all originally incorrect predictions) changes in \%, represented as ratio averages.}
\label{tab:corruption_cf_transitions}
    \vspace{-5mm}
\end{table}
\endgroup

\myparagraphwithspace{\textbf{Class transitions: False/Correct.}} We first focus on the discrepancy between seen/unseen classes. Figure \ref{fig:corruption_graph} shows that on AWA2, unseen classes are affected more severely but on SUN and CUB, both are affected approximately the same. Following the practice of \cite{yucel2020deep}, we investigate this by analyzing the class transitions for each corruption category. We calculate the following; out of all (originally) correctly predicted samples, what percentage have transitioned to false and out of all (originally) falsely predicted samples, what percentage have transitioned to correct to other false classes? The results are shown in Table \ref{tab:corruption_cf_transitions}.

We see that the CF transitions correlate with the results of Figure \ref{fig:corruption_graph}; stronger categories induce higher CF transitions. Furthermore, the stronger categories induce higher FF transitions, which is another indication of stronger corruptions inducing more class transitions. FC transitions are higher for \textit{digital} corruptions (up to 16\%), which has the lowest CF transitions (\ie 30\% on AWA2 for seen classes), which explains why it is the worst performing category.

Comparing seen and unseen class performance provides key insights. First, CF transitions are higher for unseen classes, especially on AWA2 where unseen CF ratios are nearly the double of the seen CF ratios. Furthermore, unseen FC transitions are lower compared to seen classes, except on SUN where they are similar. High CF and low FC tells us that unseen classes are affected disproportionately. We note that this is also the case for adversarial attacks, although the absolute accuracy values~\cite{yucel2020deep} shows otherwise for adversaries due to the \textit{pseudo-robustness effect}. A similar situation happens here; Figure \ref{fig:corruption_graph} shows unseen/seen discrepancy visibly only for AWA2. On CUB and SUN, the discrepancy is not as clear. Unlike the adversarial case of \cite{yucel2020deep}, however, the level of discrepancy between seen/unseen classes correlate with the absolute reduction results of Table \ref{tab:corruption_cf_transitions}; on CUB and SUN, CF/FC difference between unseen/seen classes is not prominent therefore seen/unseen absolute accuracies are close, whereas CF/FC difference between unseen/seen classes is prominent and seen/unseen absolute accuracies are quite different. Finally, we observe high FC values for seen classes; this potentially means that the correct prediction has a high softmax probability but not high enough to be correct prediction (\ie model predicts incorrectly, but correct label has high confidence), therefore any \textit{push} enforced to the image in the decision boundaries is likely to go to correct class, resulting into high FC transitions for seen classes.

\myparagraphwithspace{\textbf{Class transitions: Seen/Unseen.}} Next, we analyse the transitions from a seen/unseen class perspective. Following the practice of \cite{yucel2020deep}, we calculate the following for all samples and average it for seen and unseen classes; out of \textit{all} changed samples, what percent went to a seen or an unseen class. The results are given in Table \ref{tab:corruption_seen_unseen_transitions}. Comparing different categories shows us minimal differences, suggesting that the type of corruption has minimal role in the transitions. We also see that originally seen classes are more likely to transition to another seen class, compared to originally unseen classes. Regardless of the dataset and the corruption category, transitions happen overwhelmingly towards the seen classes. These trends are similar to the findings of \cite{yucel2020deep} and we believe the underlying reasons are the same; the discrepancy between number of seen/unseen classes in the datasets, as well as the bias of the model towards seen classes, play a role in dictating the transitions.

\begingroup
\setlength{\tabcolsep}{8pt} 
\renewcommand{\arraystretch}{1} 
\begin{table}[!t]
\resizebox{\textwidth}{!}{%
\begin{tabular}{cccclcccclccl}
\multicolumn{1}{l}{} & \multicolumn{12}{c}{\textbf{Generalized Zero Shot}}                                                                                                                                                                                                                                     \\
\multicolumn{1}{l}{} & \multicolumn{4}{c|}{C}                                                                             & \multicolumn{4}{c|}{S}                                                                             & \multicolumn{4}{c}{A}                                                         \\ \hline
Class Transition     & \multicolumn{1}{l}{UU} & \multicolumn{1}{l}{US} & \multicolumn{1}{l}{SU} & \multicolumn{1}{l|}{SS} & \multicolumn{1}{l}{UU} & \multicolumn{1}{l}{US} & \multicolumn{1}{l}{SU} & \multicolumn{1}{l|}{SS} & UU                     & \multicolumn{1}{l}{US} & \multicolumn{1}{l}{SU} & SS \\ \hline
$Noise$             & 26                     & 74                     & 23                     & \multicolumn{1}{l|}{77} & 14                     & 86                     & 12                     & \multicolumn{1}{c|}{88} & \multicolumn{1}{c}{8} & 92                    & 4                     & 96 \\

$Digital$             & 23                     & 77                     & 17                     & \multicolumn{1}{l|}{83} &16                     & 84                     & 12                     & \multicolumn{1}{c|}{88} & 12                     & 88                     & 5                     & 95 \\ \hline

$Weather$             & 25                    & 75                     & 20                     & \multicolumn{1}{l|}{80} & 16                     & 84                     & 10                     & \multicolumn{1}{c|}{90} & 12                     & 88                     & 5                     & 95 \\

$Blur$             & 22                     & 78                     & 17                     & \multicolumn{1}{l|}{83} & 14                     & 86                     & 10                     & \multicolumn{1}{c|}{90} & 10                     & 90                     & 7                     & 93 \\ \hline
\end{tabular}%
}
\caption{Corruption-induced, class transition averages (in \%) for different corruption categories. UU, US, SU and SS are unseen-to-unseen, unseen-to-seen, seen-to-unseen and seen-to-seen transitions, respectively.}
\label{tab:corruption_seen_unseen_transitions}
    \vspace{-5mm}
\end{table}
\endgroup

The majority of the adversarial robustness analysis in \cite{yucel2020deep} is driven by the fact that the ALE model is \textit{weak}; it is not really accurate. When subjected to perturbations, there are common occurrences of attacked test sets having higher accuracies. In the case of corruptions, this happens only once; the brightness corruption introduces less than 0.1\% increase in ZSL accuracy on AWA2. Since this is a negligible increase for an outlier (1 out of 75 corruptions), it is plausible to say that weakness of the model does not play a part under the presence of corruptions. 

\myparagraphwithspace{\textbf{Adverse effect of defenses.}} As shown in Figures \ref{fig:combined_1} to \ref{fig:combined_2}, the majority of the defense mechanisms do not work well. Their performance results into worsening of the results, not just into \textit{failure to recover}. In \cite{yucel2020deep}, results show the adverse effects of defenses (under adversarial attacks) had high correlation with \textit{FCF} cases (\ie original false prediction corrected by the attack, and then recovered to original prediction by the defense). In contrast, here we see that defenses do not work, because they \textit{do not work}. Since there are virtually no cases where a corruption increases the accuracy, we observe that \textit{FCF} cases do not exist, therefore the adverse effect of defenses are not tied to the \textit{weakness} of the model. 

\myparagraphwithspace{\textbf{Corrupting only the correct predictions.}} We corrupt only the originally correct predictions and then defend them; the motivation here is to decouple the effects of model weakness and ZSL-specific trends. The results here validate our previous claim; the weakness of the model does not play a part. We see the same trends in all of our results; defenses which did not work \textit{still} do not work. The only case of corruptions increasing the accuracy went away, but this is an isolated case which has no effect on the overall trends. All the findings explained in Section \ref{corruption_results} still hold; ZSL/GZSL, unseen/seen discrepancies and corruption/defense performances are the same.

\myparagraphwithspace{\textbf{ZSL vs GZSL.}} The average performance of the model in ZSL and GZSL settings show that harmonic scores are consistently more affected than ZSL accuracies. This can be credited to nature of both settings; in ZSL, all test classes are  balanced since \textit{none} of them are seen during training. In GZSL, class imbalance  takes its toll and severely impacts unseen classes, which in turn results into lower harmonic scores. For defenses that do not work, we do not have a conclusive answer as to whether ZSL or GZSL scores are recovered more effectively. For defenses that work, GZSL accuracies are recovered whereas ZSL accuracies fail to recover generally. We credit this to several factors; first, GZSL methods are severely impacted already, so there is more \textit{space} for recovery. Second, working defense methods produce models with higher GZSL accuracies, which results into better robustness.

\myparagraphwithspace{\textbf{Dataset characteristics.}} SUN and CUB both have high number of classes with few per-class sample count. AWA2, in contrast, has few number of classes with high per-class sample count. We see the effects of this in our results; AWA2 is the least affected dataset, especially in ZSL (\ie reduction of 23\% for digital category). Seen/unseen class discrepancy is most visible on AWA2, possibly due to the class imbalance further exacerbated with high per-class sample count. This is also reflected by quite low CF transitions. Highest FC transitions happen for AWA2 trained model as well, suggesting AWA2 trained model outputs multiple confident predictions. Furthermore, highest transitions to seen classes also show AWA2 to be extremely biased. Despite the fact that AWA2 is the least affected one under corruptions, it enjoys the highest recoveries when defended (\ie 50\% recovery in noise category). When defended with preprocessing defenses (\ie spatial smoothing and TVM), we observe SUN gets the best returns followed by CUB. Apart from that, SUN and CUB trained models share similar characteristics in our results.

\subsection{Comparing LFGAA and ALE}

\myparagraphwithspace{\textbf{ALE vs LFGAA.}} The results of Figures \ref{fig:corruption_graph} and \ref{fig:corruption_graph_lfgaa} show that LFGAA is more robust than ALE, especially for \textit{LFGAA+Hybrid}. We credit the improved robustness performance of LFGAA to the afore mentioned factors that make LFGAA a more accurate model in the first place.

\myparagraphwithspace{\textbf{Transductive vs Inductive.}}  The results of Figure \ref{fig:corruption_graph_lfgaa} show that \textit{LFGAA+Hybrid} does a better job then \textit{LFGAA+SA}. This is surprising because \textit{LFGAA+SA} utilizes more data, which is shown to improve robustness \cite{schmidt2018adversarially}. However, we believe that the corruptions may be harming the self-adaptation process, which results into less effective prototypes.

\myparagraphwithspace{\textbf{Defense Performance for ALE and LFGAA.}} When we compare Figures \ref{fig:combined_2}, \ref{fig:transductive_defenses} and \ref{fig:inductive_defenses}, we see that the defenses work approximately the same for ALE and \textit{LFGAA+SA}. There are some differences in performances across different datasets, but in average AugMix and ANT provides visible improvements. When we compare \textit{LFGAA+Hybrid} and ALE, we see that ALE gets better results with defenses.

\myparagraphwithspace{\textbf{Comparing the trends for ALE and LFGAA.}} We see that the trends presented for ALE in Section \ref{corruption_analyses} apply to LFGAA as well; ZSL results are less affected than GZSL, unseen classes go through a more severe degradation and AWA2-trained models are more robust than others.

\section{Comparing adversarial and corruption robustness}
\label{final_words}

We now combine our observations with our prior results on adversarial robustness from \cite{yucel2020deep} to conclude our analyses. 

\myparagraphwithspace{\textbf{Model strength.}} The original accuracy values of ALE have an impact on adversaries \cite{yucel2020deep}, whereas it does not have an effect on corruptions. Adversarial attacks try to cross decision boundaries in an \textit{efficient} manner (\ie imperceptible, minimum perturbation) whereas corruptions have no such characteristics, as can be seen from the fact that Table \ref{tab:corruption_cf_transitions} has lower FC than the corresponding results of \cite{yucel2020deep}. Adversaries generally cross to the closest decision boundary, which in many cases result into \textit{correcting} an originally incorrect prediction (\ie upward accuracy spikes). Defenses usually recover the originally false predictions (\ie adverse effects of defenses). We see neither of these for corruptions.

\myparagraphwithspace{\textbf{Unseen/seen discrepancy.}} We see that unseen classes are affected more severely. This applies to both corruption and adversarial robustness, suggesting class imbalance affects them both. However, due to model weakness, this effect is essentially \textit{masked} for adversarial attacks~\cite{yucel2020deep}, \ie unseen classes go through less degradation than seen classes, which we call the \textit{pseudo-robustness effect}. This effect is not observed for corruptions.

\myparagraphwithspace{\textbf{GZSL and ZSL.}} In corruption experiments, GZSL performance is affected more severely whereas for adversarial attacks, there is no conclusive answer \cite{yucel2020deep}. In both cases, working defenses improve GZSL performance more than ZSL.

\myparagraphwithspace{\textbf{Datasets.}} Models trained with AWA2 are the most robust ones against adversaries \cite{yucel2020deep} and corruptions, primarily owing to its high per-class sample count and potentially to its few number of classes. The robustness difference between AWA2 and other datasets is smaller against corruptions, suggesting that simply having more data may not be the ideal solution for corruptions.

\myparagraphwithspace{\textbf{Defenses.}} Our adversarial defenses do not work against corruptions. Although these work relatively well against adversaries \cite{yucel2020deep}, they fail against corruptions (except some noise and blur types). Despite the intricate design of adversaries, this suggests that their effect on images are somewhat limited; corruptions inherently cover more range of potential effects and potentially require a \textit{more} fundamental solution.

\section{Conclusion} \label{conclude}

The vulnerability of ML models against perturbations have attracted significant attention, however it has been approached from a fully supervised perspective. In this study, we perform the first comprehensive study to analyse the robustness of discriminative ZSL models against image corruptions. We subject several commonly used ZSL models to corruptions and defenses, and also create multiple corruption benchmark datasets for ZSL. Our baseline results, spanning a wide range of corruptions and defenses, show that the discriminative ZSL models are not robust, primarily due to the severe class imbalance and model weakness inherent to ZSL. Our results indicate that although some defense methods work, they fail to do so in a tangible manner, which highlights the necessity of further research. Our results also show that although these defense methods fail to work, they set new high accuracies for our ZSL models. We highlight the important differences in robustness between seen/unseen classes and ZSL/GZSL settings. Finally, we conclude our discussion jointly over the results on corruptions and adversarial attacks, and provide a larger picture of discriminative ZSL robustness, both from an adversarial and a corruption robustness perspective.

\textbf{Acknowledgments.} This work is performed as a part of the Ph.D. studies of the first author.

\bibliography{egbib}

\end{document}